\documentclass[conference]{IEEEtran}
\usepackage{times}
\usepackage{cite}
\usepackage[numbers]{natbib}
\usepackage{multicol}
\usepackage{graphicx}
\usepackage[bookmarks=true]{hyperref}
\hypersetup{
  colorlinks   = true, 
  urlcolor     = blue, 
  linkcolor    = blue, 
  citecolor   = blue 
}
\usepackage[format=plain,
            labelfont={bf,it},
            textfont=it]{caption}
\usepackage{subcaption}
\usepackage{epstopdf}
\usepackage{mathtools,amsmath,amssymb,lipsum}

\usepackage{cuted}
\newcommand{\subparagraph}{}
\usepackage[compact]{titlesec}

\begin{document}

\title{ORangE: Operational Range Estimation for Mobile Robot Exploration on a Single Discharge Cycle}

\author{\authorblockN{Kshitij Tiwari\authorrefmark{1}, Xuesu Xiao\authorrefmark{2}, Ville Kyrki\authorrefmark{1}, and Nak Young Chong\authorrefmark{3}}
\authorblockA{\authorrefmark{1}School of Electrical Engineering, Aalto University, Finland, 02150\\ Email: \{kshitij.tiwari, ville.kyrki\}@aalto.fi}
\authorblockA{\authorrefmark{2} Dept. of Comp. Sci. \& Eng., Texas A\&M University, College Station, TX 77843\\
Email: xiaoxuesu@tamu.edu}
\authorblockA{\authorrefmark{3}School of Info. Sci., JAIST, Japan, 923-1211\\ Email: nakyoung@jaist.ac.jp}}

\maketitle

\begin{abstract}
This paper presents an approach for estimating the operational range for mobile robot exploration on a single battery discharge. Deploying robots in the wild usually requires uninterrupted energy sources to maintain the robot's mobility throughout the entire mission. However, due to unknown nature of the environments, recharging is usually not an option, due to the lack of pre-installed recharging stations or other mission constraints. In these cases, the ability to model the on-board energy consumption and estimate the operational range is crucial to prevent running out of battery in the wild. To this end, this work describes our recent findings that quantitatively break down the robot's on-board energy consumption and predict the operational range to guarantee safe mission completion on a single battery discharge cycle. Two range estimators with different levels of generality and model fidelity are presented, whose performances were validated on physical robot platforms in both indoor and outdoor environments. Model performance metrics are also presented as benchmarks. 
\end{abstract}

\IEEEpeerreviewmaketitle

\section{Introduction}
Autonomous robots are becoming ubiquitous in all walks-of-life: agriculture \cite{bakker2006autonomous}, logistics \cite{cosma2004autonomous}, household \cite{fiorini2000cleaning}, defense \cite{naskar2011application}, search-and-rescue \cite{xiao2017uav}, to name but a few. Majority of the robots used in such applications, are powered by energy sources like the Li-Po battery. Depending on the end-use, it is often not possible to recharge the batteries in the form of solar energy or electric charging stations distributed over the entire workspace. While there is on-going research which focuses primarily on the optimal recharging strategies \cite{silverman2002staying,wawerla2008optimal}, the aim of this work is to highlight a more pressing concern: 
\begin{quote}
    \textit{How to ensure the robots return to base whilst avoiding complete immobilization amidst the mission?}
\end{quote}

\begin{figure}[!htbp]
\centering
\begin{subfigure}{.25\textwidth}
  \centering
  \includegraphics[scale=0.15]{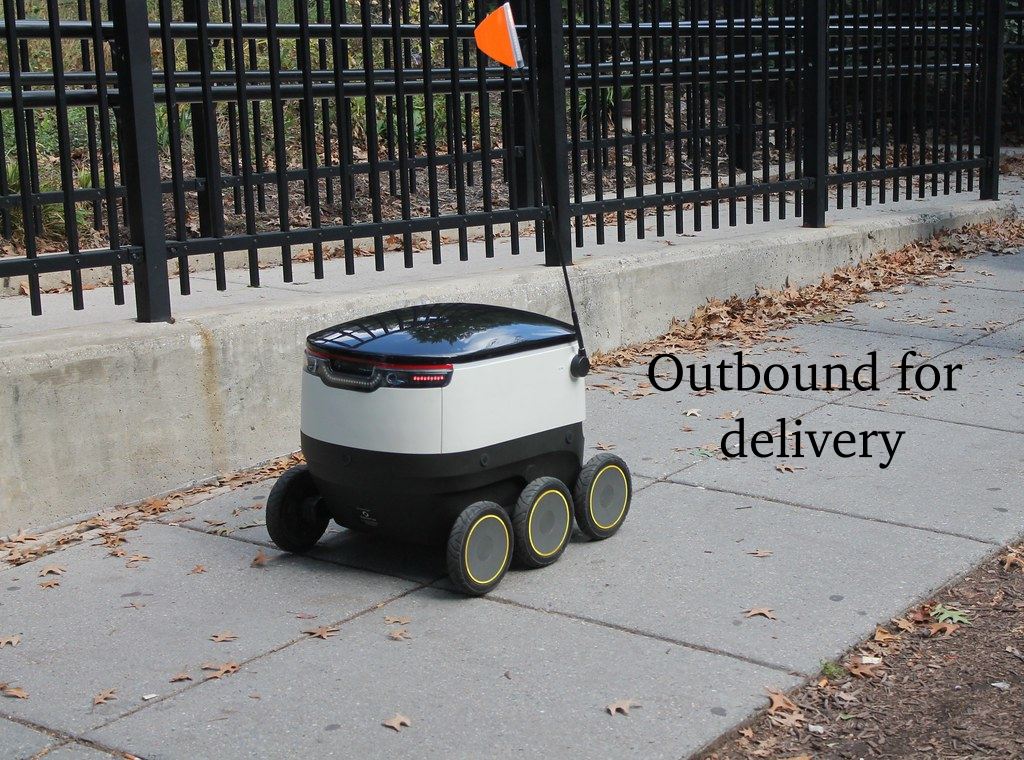}
  \caption{Delivery robot called \textit{Kiwibot}.}
  \label{subfig:delivery}
\end{subfigure}%
\begin{subfigure}{.25\textwidth}
  \centering
  \includegraphics[scale=0.145]{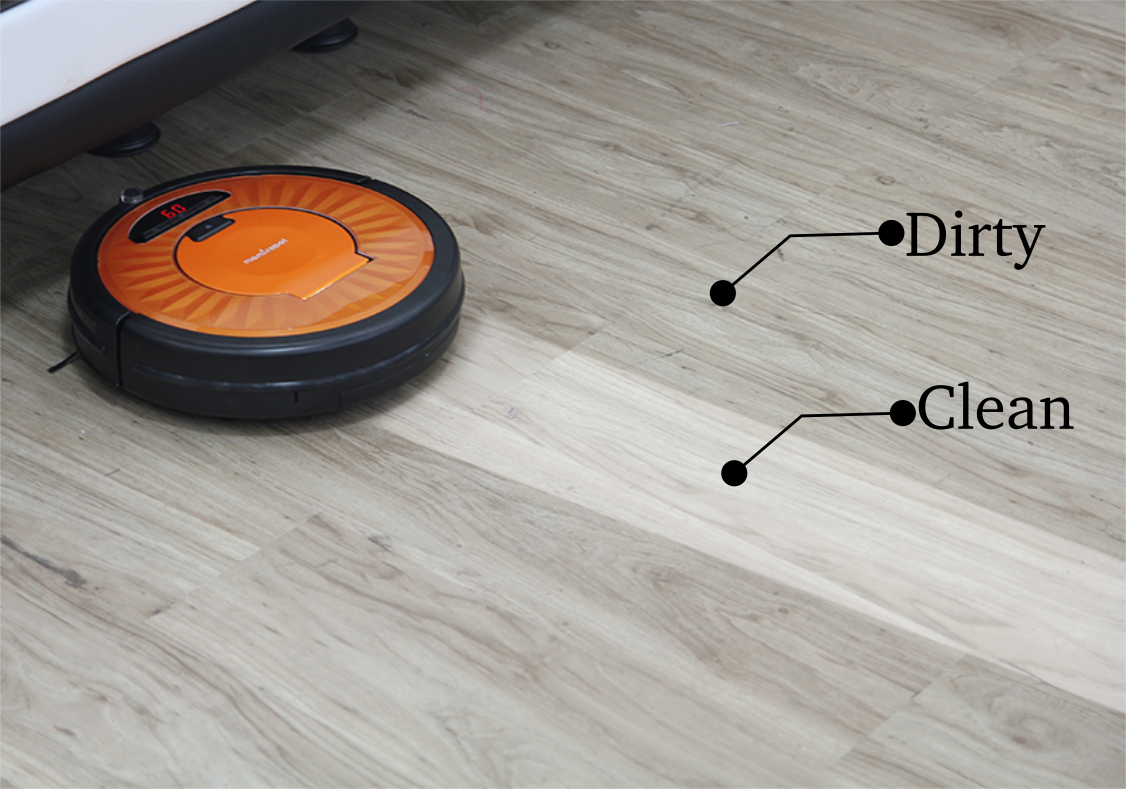}
  \caption{Vacuum cleaning robot.}
  \label{subfig:cleaning}
\end{subfigure}
\caption{Some recent use-cases of autonomous robots.}
\end{figure}

Consider a scenario where a robot is being used to deliver packages as shown in Fig.~\ref{subfig:delivery}. This application is gaining traction amongst college communities in America and is being harnessed by start-ups like Kiwibot, who use robots for modernizing food delivery. A more common sight for a tech-savy household would be to have a vacuum cleaning robot like the one shown in Fig.~\ref{subfig:cleaning} to automate mundane tasks, like cleaning the floors. Whilst these robots are modernizing the way of life, the mission will fail when the robot runs out of battery. This would mean either a delivery robot may never reach its destination or return to base, or a vacuum cleaning robot may only be able to partially clean the floor. If a robot is completely immobilized amidst its task, it requires extra manpower to retrieve the stranded robot, adding more workload for the human supervisors. 

The aim of this work then is to formalize this failure in terms of the \textit{operational range estimation} for a single discharge cycle. Unforeseen hardware failures aside, this work summarizes our recent findings in the domain of \textit{operational range estimation}, which allows for \textit{offline} or \textit{online} mission planning with homing constraints. To this end, a simplified \textit{operational range estimation (ORangE)} framework is first presented which is then extended to be applicable for a variety of robot platforms.

In what follows, two schools of mission characterization are described: the endurance and energy estimation approach is discussed, followed by our novel operational range estimation approach. Empirical results are also presented in support of our model to validate the model fidelity. This work is a brief overview of our findings that were recently published in \cite{tiwari2018estimating} and \cite{tiwari2019unified}.

\section{Endurance \& Energy Estimation}
A body of research focuses on the endurance and energy estimation approaches to characterize the mission requirements. On the one hand, endurance estimation is primarily concerned with vehicles which constantly consume kinetic energy even for hovering in place, such as fixed-winged or rotary-winged drones as shown in \cite{traub_2013} and \cite{gatti2015maximum}, respectively. On the other hand, energy estimation has primarily been shown for unmanned ground vehicles, which do not require energy to maintain a stationary workspace configuration, like the PackBot \cite{abdilla2015power}. Thus, both estimators have contrasting approaches when it comes to modeling the stationary kinematic energy consumption. Also, in either case, only the kinetic energy dispensed in Cartesian space is considered, whilst ignoring the energy used for other robotic functions aside from locomotion. 

In case of endurance estimation, existing works mainly focuses on hovering energy consumption for rotor-crafts and soaring for fixed wings in wind tunnel. However,  simply hovering and soaring does not represent the full maneuvering capacity of either kind of vehicle. Therefore, the endurance could be largely over-estimated. Similarly, in case of energy estimation, researchers focus on simplified and pre-meditated trajectories as described in \cite{sadrpour2013mission}. 

\section{Operational Range Estimation (ORangE)}
Compared to the body of research described above, we pose the mission characterization as a problem where the robots' resources must dictate how the mission progresses. For this, we present two of our proposed approaches that allow operational range estimation (ORangE) for a variety of robot platforms.

\subsection{Simplified Range Estimation Framework}
\begin{figure}[!htbp]
\centering
\begin{subfigure}{.25\textwidth}
  \centering
  \includegraphics[trim=0cm 0cm 0cm 0cm,clip=true,scale = 0.14]{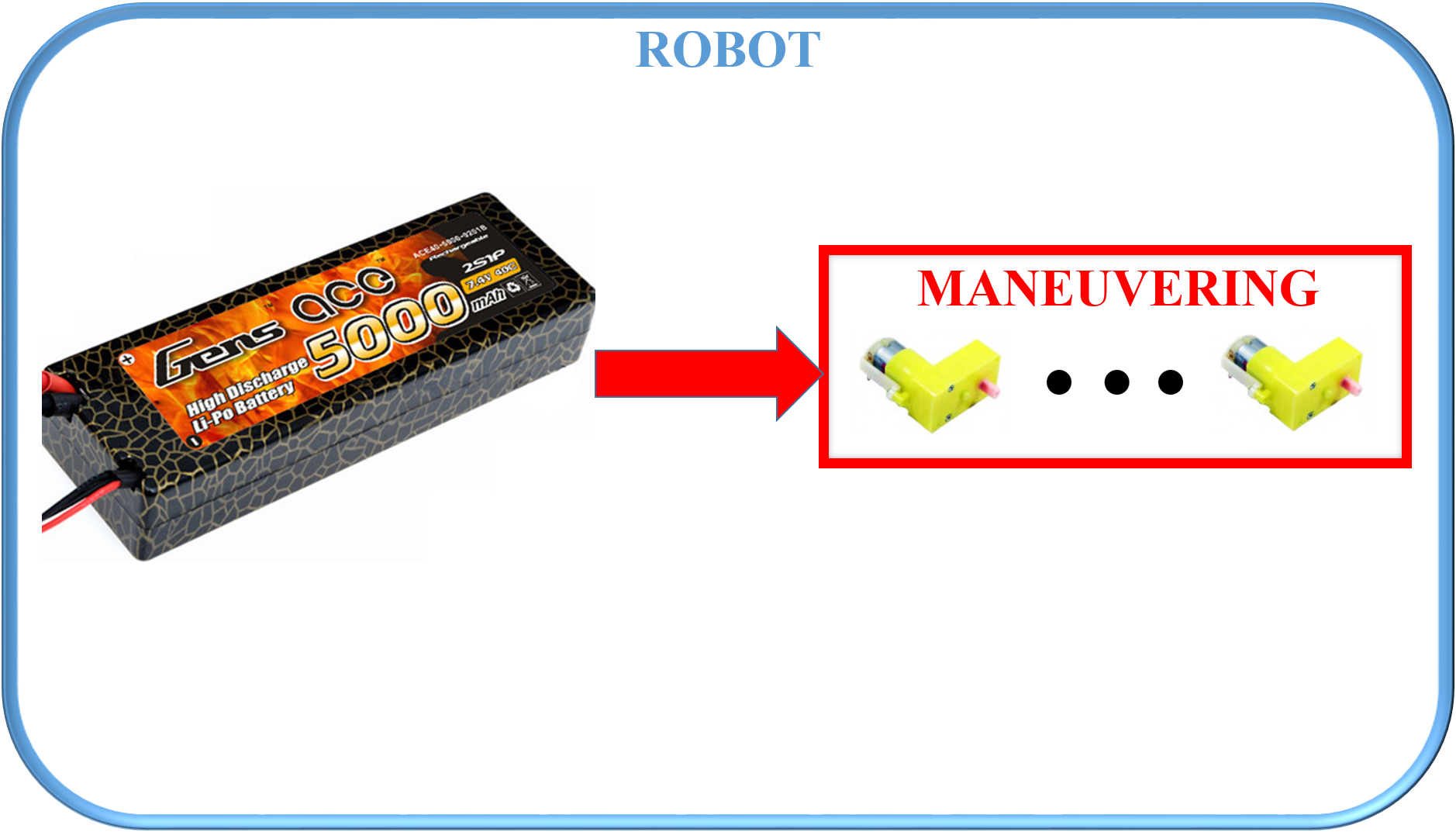}
  \caption{Idealistic model.} \label{subfig:BatteryIdeal}
\end{subfigure}%
\begin{subfigure}{.25\textwidth}
  \centering
  \includegraphics[trim=0cm 0cm 0cm 0cm,clip=true,scale = 0.14]{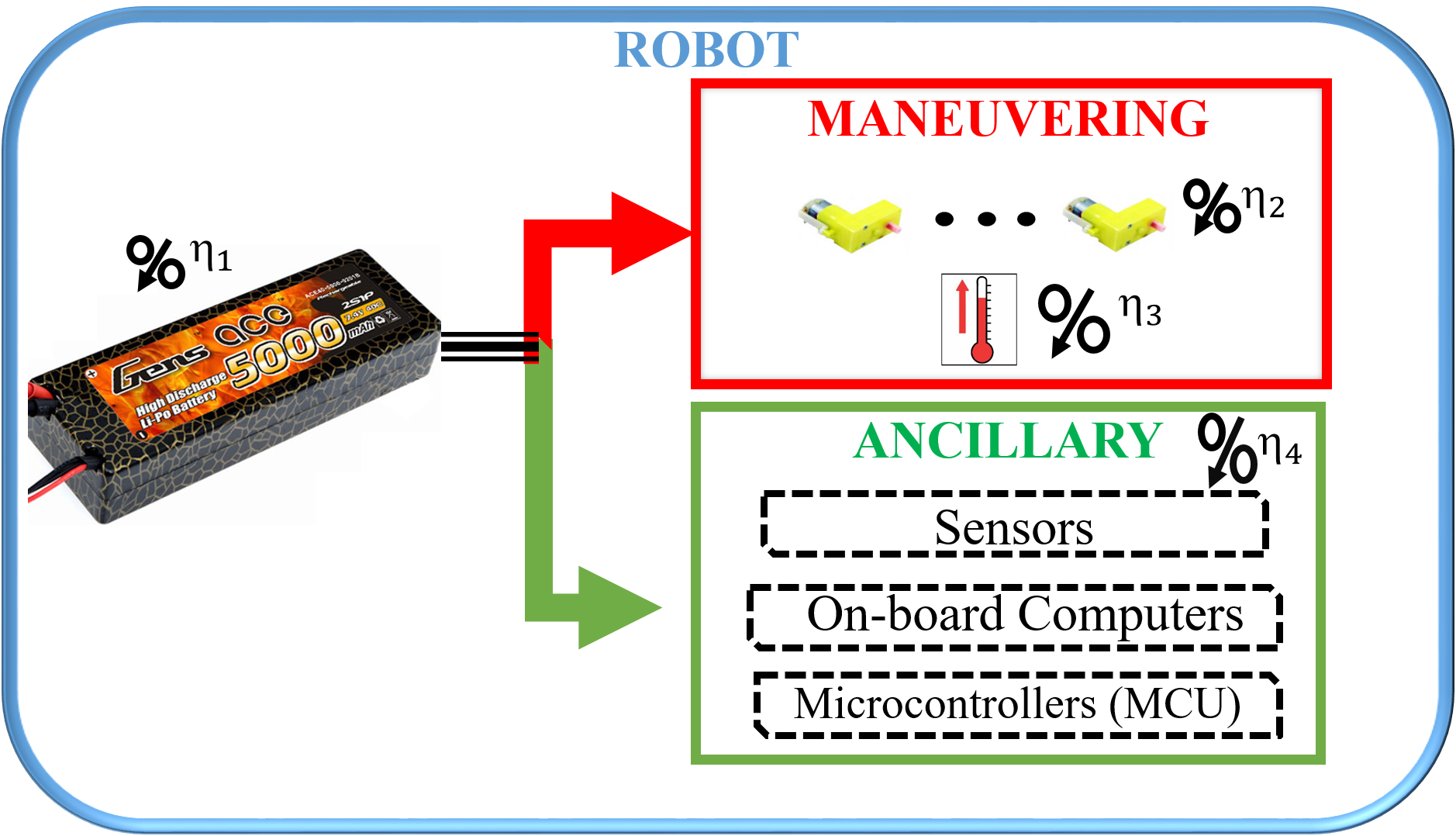}
  \caption{Realistic model.}
  \label{subfig:BatteryReal}
\end{subfigure}
\caption{In Fig.~\ref{subfig:BatteryIdeal}, we show an idealistic battery dissemination model, where all the energy stored in the battery is used as it is for performing maneuvers. In Fig~\ref{subfig:BatteryReal}, we present a realistic model where we account for battery losses $(\eta_1)$, maneuvering losses $(\eta_2,\eta_3)$ and ancillary losses $(\eta_4)$.}
    \label{fig:BatteryModel}
\end{figure}

For robots venturing into unknown environments for exploration purposes, it would be ideal that all the energy carried on-board in the form of battery is utilized in locomotion, such as maneuvering and propulsion from the motion actuators (Fig. \ref{subfig:BatteryIdeal}). In such a case, the exploration can span longer distances and wider coverage of the unknown region. However, energy carried on-board has to be dispensed for a variety of purposes, including sensing, computation, communications, friction, heating, \textit{etc.}  \cite{xiao2014william, tiwari2018estimating} (Fig. \ref{subfig:BatteryReal}). Due to their contradicting effects on achieving better operational range, energy consumed by such consumers is in fact termed as \textit{losses} which are described next.

\subsubsection{Losses}
Energy losses could occur due to a variety of reasons. Some are inevitable due to physical and chemical properties. Others are introduced deliberately by the designer for the mission. Due to the fact that all these sources impede the robot from venturing further, we generalize them into the following four groups: 
\begin{itemize}
\item \textbf{Battery charge storage loss $(\eta_1)$:} refers to the battery self-discharge characteristics. Even without any load attached, the battery tends to suffer self-discharge thereby reducing the net amount of energy available for a mission.
\item \textbf{Drive motor losses $(\eta_2)$:} owing to internal friction along with actuation losses.
\item \textbf{Mechanical losses $(\eta_3)$:} refers to power train losses like friction in transmission, damping from lubricants, \textit{etc.}
\item \textbf{Ancillary losses $(\eta_4)$:} accounts for heat losses incurred by sensors, motor drivers, micro-controllers, \textit{etc.}  
\end{itemize}
So, the overall system efficiency can be summarized as $\Omega \overset{\Delta}{=} \Pi_{i=1}^4\neg\eta_i$. In order to obtain these losses quantitatively for a particular robot, wheels-up tests were performed to eliminate the useful work done by the robot to overcome resistance from the environment during locomotion. The isolated energy consumption is only contributed by those internal losses \cite{tiwari2018estimating}. Next, a simplified range estimation model for indoor exploration is presented.

\subsubsection{Range Estimation Model}
For a robot weighing $m_R~g$ and traversing on an elevated surface with an elevation of $\theta$, the normal force $(N)$ supporting the robot and the traction force propelling it are: 
\begin{align}
\begin{split}
N &= m_R ~g \cos\theta \,.\\
Traction&= C_{rr}~N + cv^2 +m_R ~g \sin\theta \,.
\end{split}
\label{eqn:fbd}
\end{align}

In Eq.~\eqref{eqn:fbd}, $C_{rr}~N$ represents friction, $cv^2$ is the aerial draf force, and $m_R ~g \sin\theta$ is the weight component impeding the motion. Therefore, the energy needed for displacing the robot by an amount $d$ on a graded plane (referred to as the \textcolor{red}{\textbf{Maneuvering}} Energy (ME)) can be given by:

\begin{align}
\begin{split}
ME &= Traction\times d\,,\\
&= (C_{rr}~N + cv^2 + m_R ~g \sin\theta)d\,,\\
& = (C_{rr}~m_R ~g \cos\theta + cv^2 + m_R ~g \sin\theta)d\,.
\end{split}
\label{eqn:MEElevated}
\end{align}

Eq.~\eqref{eqn:MEElevated} is the mechanical energy necessary to displace the robot. However, the realistic energy losses are not considered yet. Fig.~\ref{subfig:BatteryReal} shows the other consumers of energy from the battery which are referred to as the \textcolor{green}{\textbf{ancillary}} consumers.

With regards to the battery charge storage loss ($\eta_1$), an exponential decay function is suggested to represent the reduced battery capacity due to aging $(t)$ and recharging cycles $(C)$ as:
\begin{align}
\hat{E} = E_O \exp^{-(k_1C+k_2t)} 
\label{eqn:EnergyDecay}
\end{align}
where $k_1,k_2$ are constant coefficients and $E_O$ represents the rated energy.

The total available energy from the battery is composed of ancillary energy (AE) and traversal energy (TE):

\resizebox{0.99\linewidth}{!}{
\begin{minipage}{1.16\linewidth}
\begin{align}
\begin{split}
\hspace*{-1cm} \hat{E}&= AE + TE \,, \\
 &= Ancillary ~Power \times time + \dfrac{ME}{{}^{r}\Omega_{Man}} \,,\\
 &= P_{Anc}\times\dfrac{d}{vD} + \dfrac{(C_{rr}~m_R ~g \cos\theta + cv^2 + m_R ~g \sin\theta)d}{{}^{r}\Omega_{Man}}\,,\\
  &= d \times \left\lbrace \dfrac{P_{anc}}{vD} + \dfrac{(C_{rr}~m_R ~g \cos\theta + cv^2 + m_R ~g \sin\theta)}{{}^{r}\Omega_{Man}} \right\rbrace\,.
\end{split}
\label{eqn:totalEnergy}
\end{align}
\end{minipage}
}

where $D$ is \emph{duty cycle} which is the proportion of net mission time that the robot was actually mobile and $vD$ represents the average velocity throughout the mission. While $D$ accounts for the fact that the robot may occasionally need to stop and process the data acquired, ${}^{r}\Omega_{Man} = \neg \Pi_{i=2}^{3}\eta_i$ represents the constant maneuvering efficiency of the robot $(r)$. The simplified linear model of ancillary power is given by:

\begin{align}
P_{Anc}= \underbrace{\{s_0+ s_1 f_s\}}_\text{$P_{Sense}$}
\label{eqn:AncillaryPower_simple}
\end{align}
which defines the linear increase in power consumption as a function of the operational frequency $f_s$ for sensors like camera, laser range finders, sonars, \textit{etc.,} as defined in \cite{tiwari2018estimating}.

The maximum range $d_{max}$ is achievable at an operational velocity $(v_{opt})$, that is the maximal target velocity attainable for safe operation as determined by the human supervisor. Thus,

\begin{align}
\begin{split}
&d_{max} =\\ &\left\lbrace \dfrac{\hat{E}}{\dfrac{P_{Anc}}{v_{opt}D} + \dfrac{(C_{rr}~m_R ~g \cos\theta + cv^2 + m_R ~g \sin\theta)}{{}^{r}\Omega_{Man}}} \right\rbrace
\end{split}
\label{eqn:NewasympDist}
\end{align}

\subsection{Generalized Range Estimation Framework}
The previous section described the simplified ORangE approach which was primarily focused on unmanned ground robots operating in indoor environments with known elevation $(\theta)$. However, in most missions, especially those conducted outdoors, this may not always be readily available. Additionally, missions might involve aerial or marine robots aside from only ground robots. This section presents our generalized ORangE approach, which is suited for a variety of robot platforms and allows for both \textit{offline} and \textit{online} estimation of the operational range. For this, the revised energy dissipation model will be described first.

\subsubsection{Energy Dissipation Model}

As for the \textit{propulsive energy}, any robot $(r)$ carrying out a mission $(m)$ in an environment of choice experiences $4$ kinds of forces: 
\begin{enumerate}
\item Constant resistive force $F(r,m)$, as a function of robot $(r)$ and the mission $(m)$: \textit{e.g.,} the force acting on a robot when it is traversing in a straight line under the influence of a constant magnetic field.
\item Environment dependent force $F(x,r,m)$, which is dependent on the current position $x$: \textit{e.g.,} changing gravitational potential along with changing frictional force because of change in coefficient of friction.
\item Time dependent resistive force $F(t,r,m)$, which is a function of current time $t$: \textit{e.g.,} unforeseeable disturbances (strong wind gusts \textit{etc.}).
\item Instantaneous operational velocity dependent resistive force $F(v,r,m)$, which varies with instantaneous velocity $v$: \textit{e.g.,} aerodynamics and gyro effect.
\end{enumerate}

Thus, the (revised) net \textit{traversal energy (TE)} is given in terms of \textit{mechanical energy (ME)} based on the longitudinal dynamics model and the net mechanical efficiency $({}^{r}\Omega_{Man})$ as: 
\begin{align}
\begin{split}
&TE = \dfrac{ME}{{}^{r}\Omega_{Man}}= \dfrac{\int\limits_{Path} F_{Net}dx}{{}^{r}\Omega_{Man}} \\= &\dfrac{\int\limits_{Path} \lbrace F(r,m)+F(x,r,m)+F(t,r,m)+F(v,r,m)\rbrace dx}{{}^{r}\Omega_{Man}}
\end{split}
\label{eq:traversal_generic}
\end{align}

Then, the duration $(t)$ can be expressed as a function of position $(x)$, velocity $(v)$, mission $(m)$ and duty cycle $(D)$ as:  
\begin{align}
t = g(x,v,D,m)
\label{eq:instantaneousTime}
\end{align}

Substituting Eq.~\eqref{eq:instantaneousTime} into Eq.~\eqref{eq:traversal_generic} gives :

\begin{align}
\begin{split}
TE &= \dfrac{\lbrace F(r,m)+F(v,r,m)\rbrace d}{{}^{r}\Omega_{Man}} \\ &+\dfrac{d \int\limits_{Path} \lbrace F(x,r,m)+F(x,v,D,r,m)\rbrace dx}{d~{}^{r}\Omega_{Man}} 
\end{split}
\end{align}

As an enhancement over the ancillary power consumption model shown in in Eq.~\eqref{eqn:AncillaryPower_simple}, the generalized ancillary power model of Eq.~\eqref{eqn:AncillaryPower_generic} additionally considers the on-board computation cost and the communication overhead incurred due to data transmission as explained in \cite{tiwari2019unified}.

\begin{align}
P_{Anc}= \underbrace{\{s_0+ s_1 f_s\}}_\text{$P_{Sense}$} + \underbrace{\{P_{Comp}+P_{Comm}\}}_\text{\textcolor{black}{$P_{c}$}}
\label{eqn:AncillaryPower_generic}
\end{align}

\subsubsection{Range Estimation Model}
So, similar to Eq.~\eqref{eqn:totalEnergy}, the operational range for any robot $(r)$ can be can now be generalized as:

\begin{strip}
\begin{align}
d = \dfrac{\tilde{E}}{\dfrac{P_{Anc}}{vD} +\dfrac{\lbrace F(r,m)+F(v,r,m)\rbrace}{{}^{r}\Omega_{Man}} +\dfrac{\int\limits_{Path} \lbrace F(x,r,m)+F(\dfrac{x}{vD},r,m)\rbrace dx}{d~{}^{r}\Omega_{Man}}}
\label{eq:OperationalRangeAnyRobot}
\end{align}
\end{strip}

From Eq.~\eqref{eq:OperationalRangeAnyRobot}, it is evident that in order to estimate the operational range, we need to approximate the term $\frac{\int\limits_{Path}\lbrace F(x,r,m)+F(\dfrac{x}{vD},r,m)\rbrace dx}{d~{}^{r}\Omega_{Man}}$ and the operational range estimate would be as good as the approximation. Depending on whether this integral is approximated with real-time mission data or approximated in one-shot by the human supervisor \textit{a priori}, we get \textit{online} or \textit{offline} estimators, respectively, the details of which can be found in \cite{tiwari2019unified}.

\section{Experiments}
This section provides the empirical results of the simplified and generic range estimators to validate their respective model fidelity. \textit{First,} the robot platforms used for validation are presented followed by results from the indoor trials and finally, the outdoor experiments with two different platforms. Whilst the indoor environments are considerably safe operational settings for robots, outdoor environments present challenging operational conditions owing to unforeseen environmental disturbances like wind gusts or sudden rain showers. Thus, the proposed approaches were tested in myriad of conditions. 

\subsection{Robot Platforms}
\begin{figure}[!htbp]
\centering
\begin{subfigure}{.15\textwidth}
  \centering
  \includegraphics[width=0.95\textwidth, height=0.08\textheight]{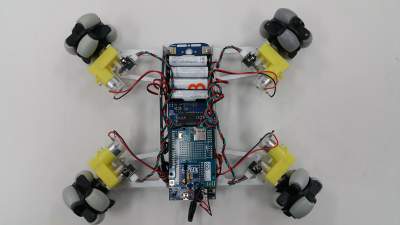}
  \caption{Rusti V$1.0$.}
\end{subfigure}%
\begin{subfigure}{.15\textwidth}
  \centering
  \includegraphics[width=0.95\textwidth, height=0.08\textheight]{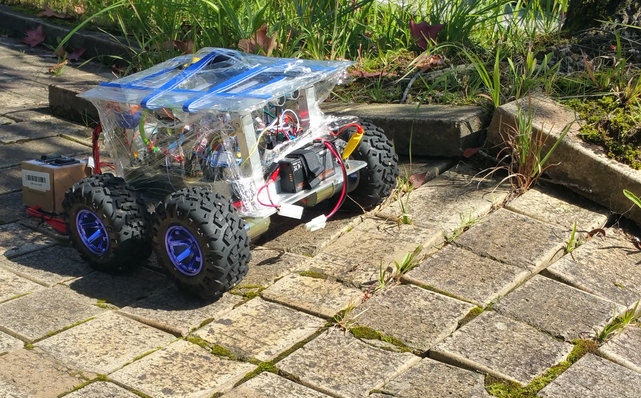}
  \caption{Rusti V$2.0$.}
\end{subfigure}%
\begin{subfigure}{.15\textwidth}
  \centering
  \includegraphics[width=0.95\textwidth, height=0.08\textheight]{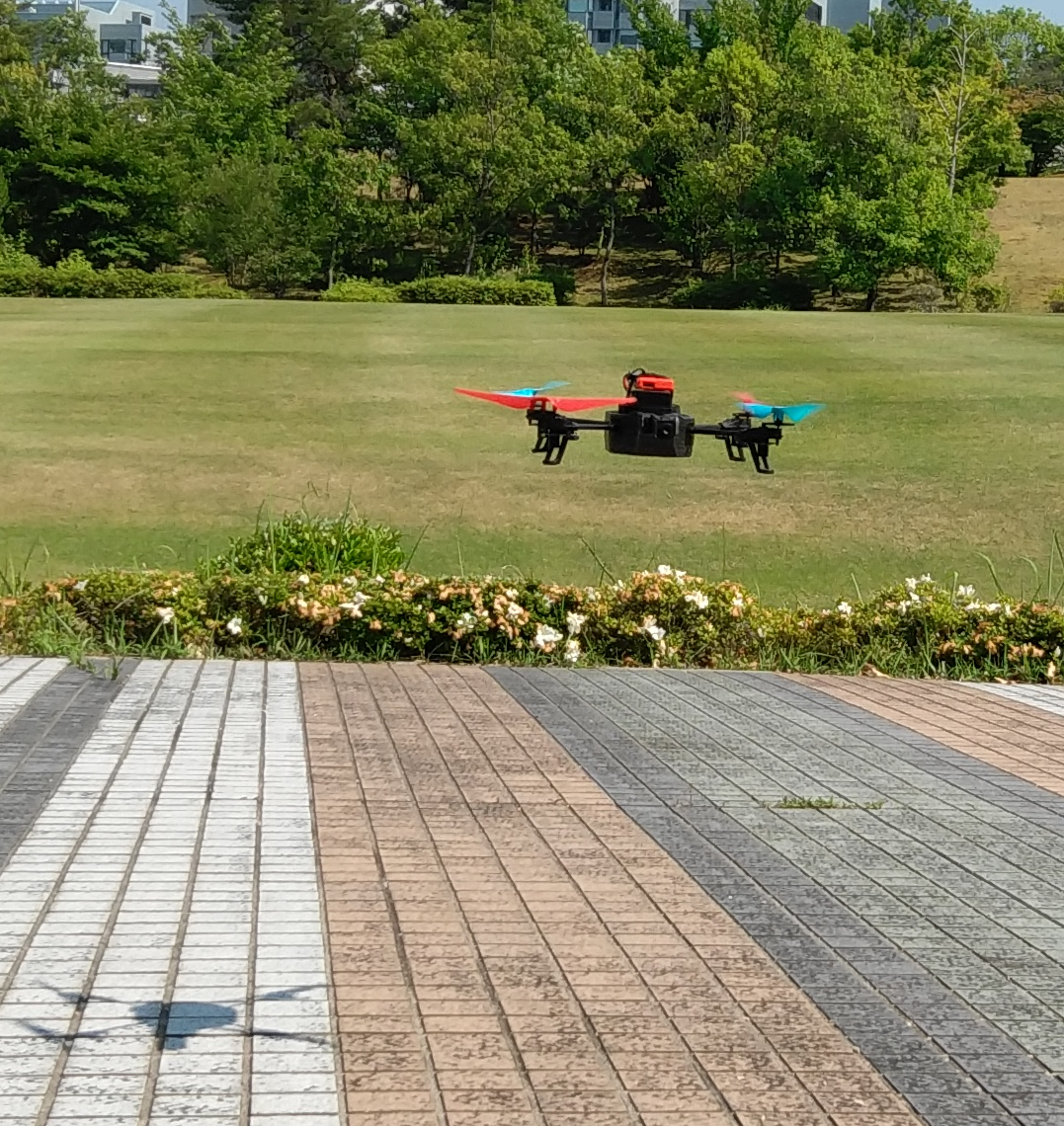}
  \caption{AR Drone $2.0$.}
\end{subfigure}
\caption{Various custom and commercial robot platforms for empirical validation of range estimators.}
\label{fig:robots}
\end{figure}

The results presented herewith were obtained using the robot platforms shown in Fig.~\ref{fig:robots}.

\subsection{Indoor Validation of Simplified ORangE with UGV}
\begin{figure}[!htbp]
\centering
\begin{subfigure}{.25\textwidth}
  \centering
  \includegraphics[trim=0cm 0cm 0cm 0cm,clip=true,scale=0.13]{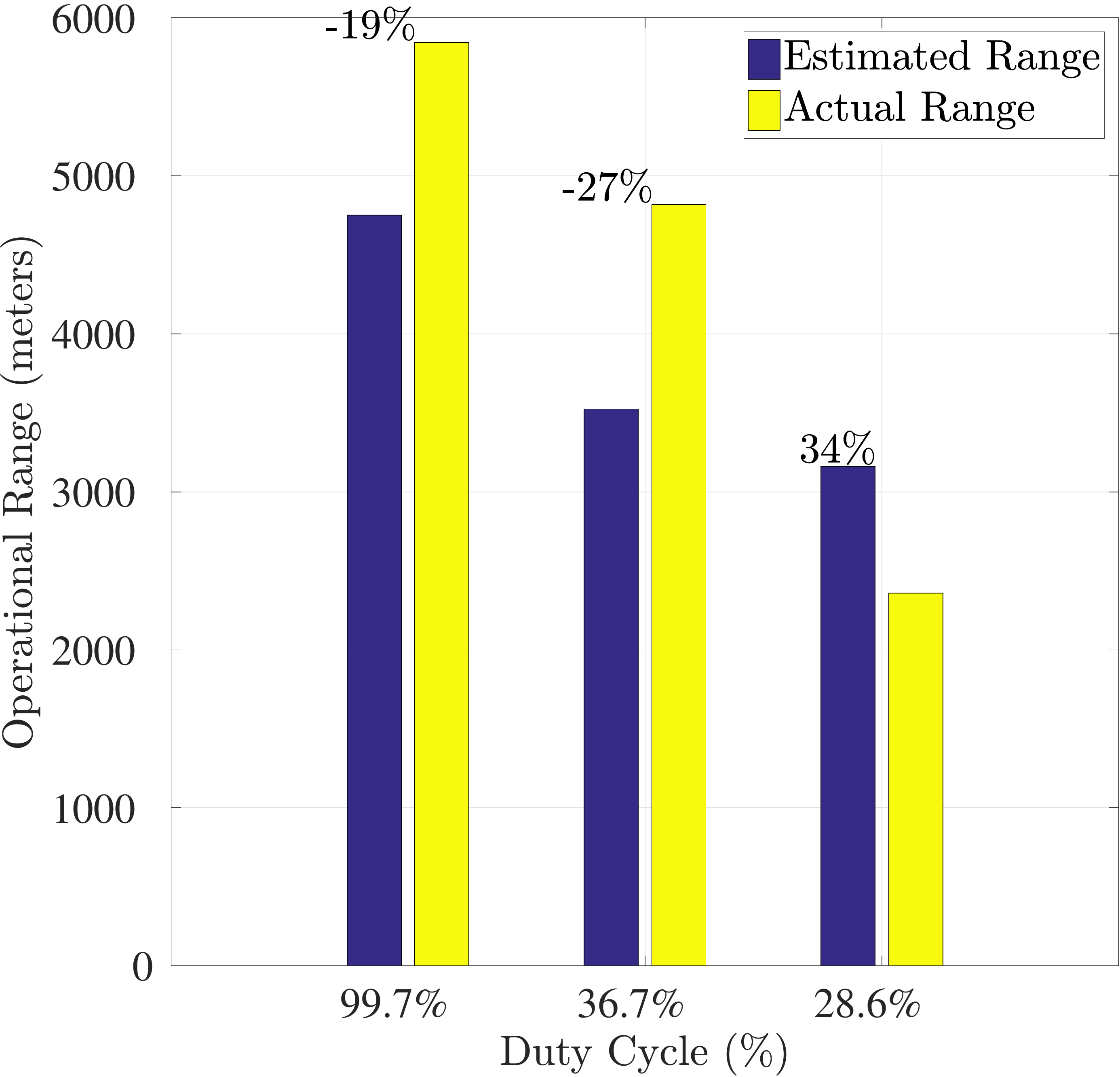}
  \caption{Flat plane.}
  \label{subfig:RangeBox}
\end{subfigure}%
\begin{subfigure}{.25\textwidth}
  \centering
  \includegraphics[trim=0cm 0cm 0cm 0cm,clip=true,scale=0.13]{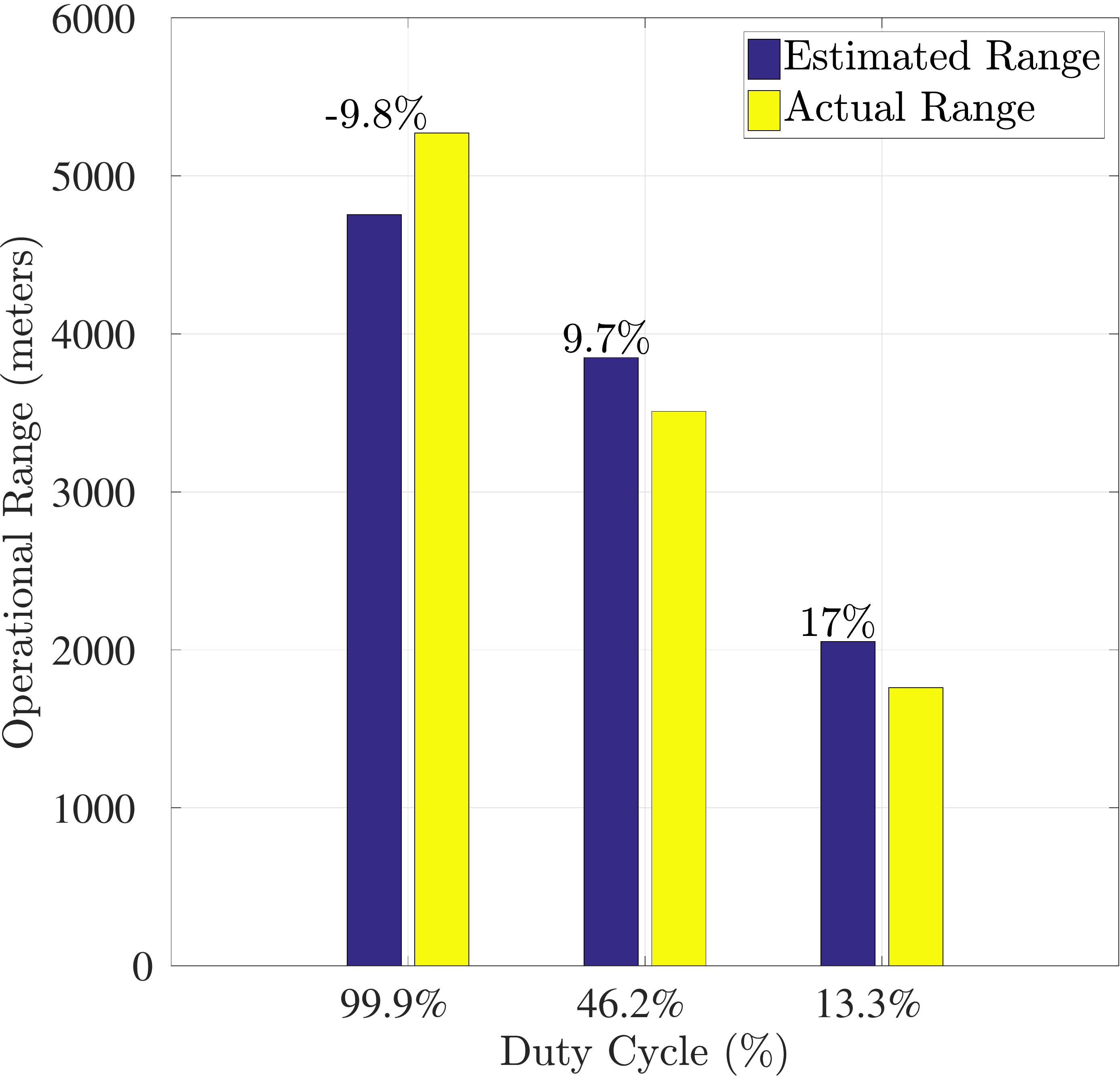}
  \caption{Graded plane.}
  \label{subfig:RangeSlope}
\end{subfigure}
\caption{Range estimation error for flat plane and graded slope experiments using Rusti V$1.0$ operating at various duty cycles.}
    \label{fig:EstimationError_rusti_v1.0}
\end{figure}

Using the Rusti V$1.0$ (shown in Fig.~\ref{fig:robots}, left) for indoor navigation scenario, the following two kinds of experiments were performed: \textit{Firstly,} the robot was made to execute a box-type trajectory on a flat floor until a fully charged battery was completely drained. \textit{Secondly,} the robot was made to execute an oscillating trajectory on a mildly graded plane. The results obtained are shown in Fig.~\ref{fig:EstimationError_rusti_v1.0} \cite{tiwari2018estimating}. As is evident, the model simplifications and noisy sensors rendered the model to achieve an accuracy ranging from $66\%\sim 91\%$.

\subsection{Outdoor Validation of Generalized ORangE}
For the validation of the generalized ORangE approach, several outdoor trails were performed with custom designed Rusti V$2.0$ and commercial AR Drone platforms as shown in Fig.~\ref{fig:robots}. Some of the results are illustrated below.

\subsubsection{Unmanned Ground Robot}
\begin{figure}[!htbp]
\centering
\begin{subfigure}{.15\textwidth}
  \centering
  \includegraphics[trim=0cm 1cm 0cm 0cm,clip=true,scale=0.18]{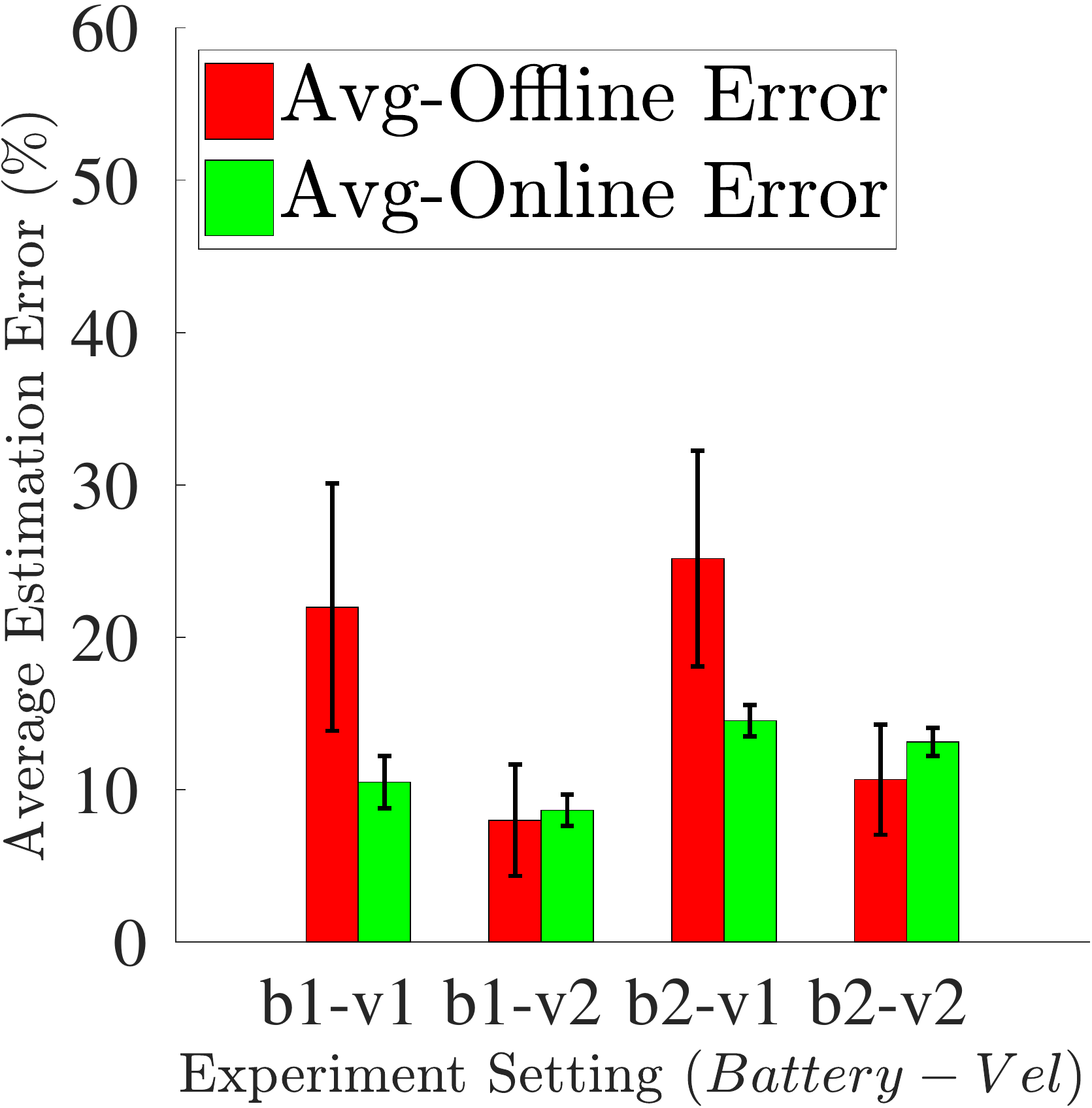}
  \caption{Grass trials.}
  \label{subfig:ErrorGrass}
\end{subfigure}%
\begin{subfigure}{.15\textwidth}
  \centering
  \includegraphics[trim=0cm 1cm 0cm 0cm,clip=true,scale=0.18]{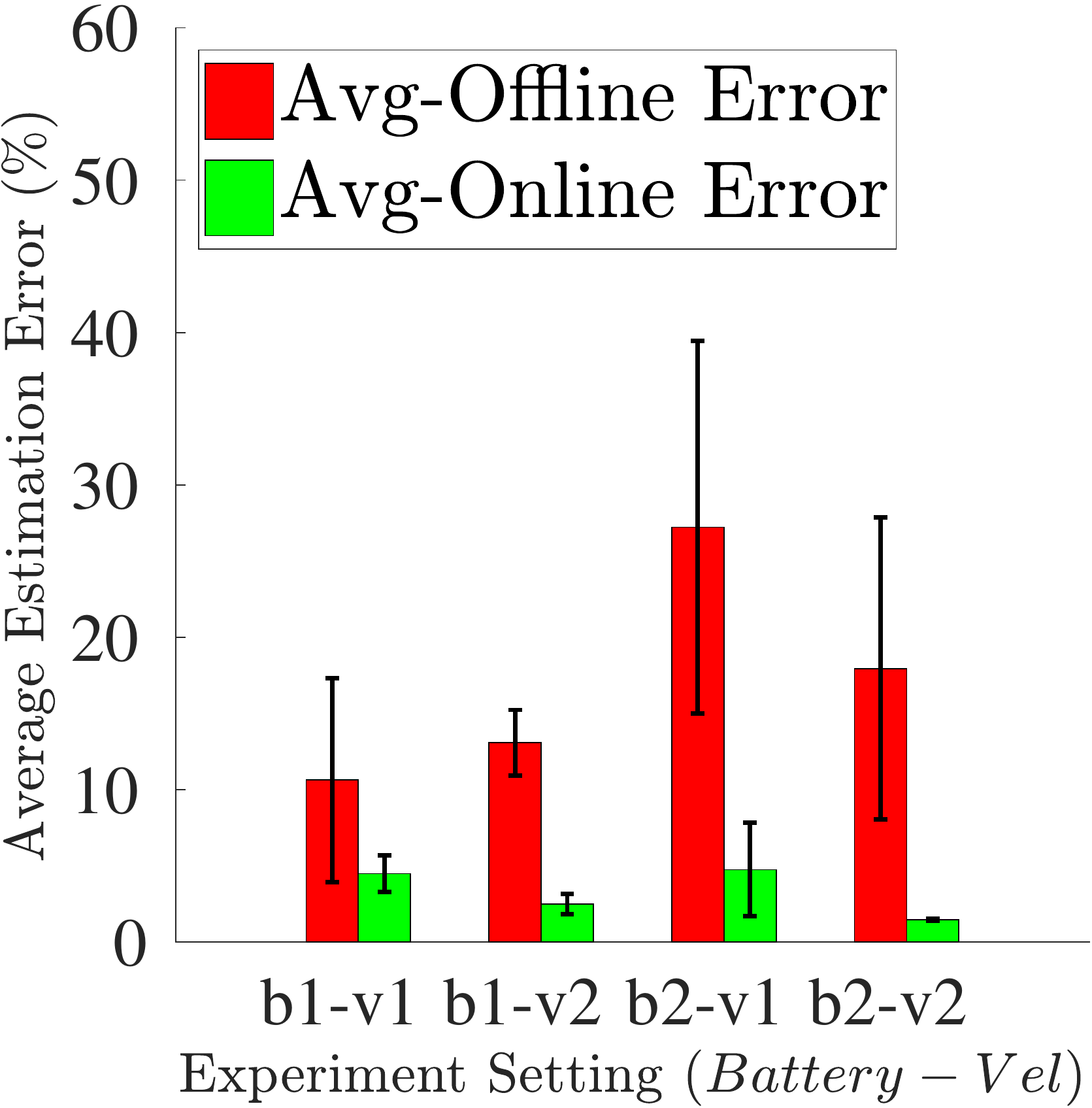}
  \caption{Asphalt trials.}
  \label{subfig:ErrorAsphalt}
\end{subfigure}%
\begin{subfigure}{.15\textwidth}
  \centering
  \includegraphics[trim=0cm 1cm 0cm 0cm,clip=true,scale=0.18]{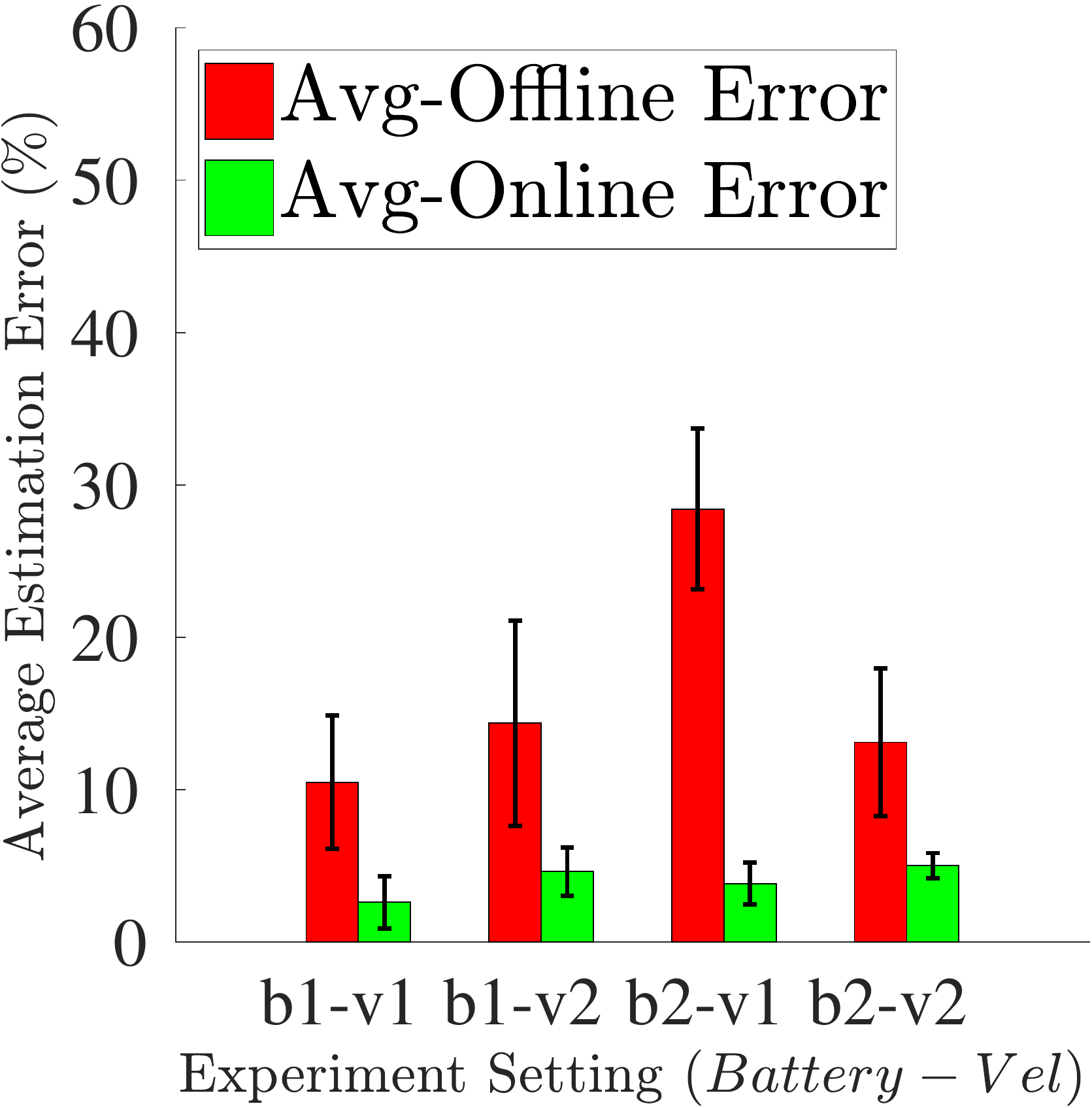}
  \caption{Tile trials.}
  \label{subfig:ErrorTile}
\end{subfigure}
\caption{Range estimation error for Rusti while traversing on grass, asphalt and tiles, respectively. Here $b1,b2$ refer to the $1500$ mAh and $2200$ mAh batteries and $v1,v2$ refers to $0.544,0.952$ m/sec velocities, respectively.}
\label{fig:RangeErrorRusti}
\end{figure}

In order to validate the generalized ORangE for unmanned ground robots, several experiments were performed on grass, asphalt, and tiled surfaces using the Rusti V$2.0$. The results hence obtained are show in Fig.~\ref{fig:RangeErrorRusti}.

\subsubsection{Unmanned Aerial Vehicle}
\begin{figure}[!htbp]
 \centering
 \includegraphics[scale=0.23]{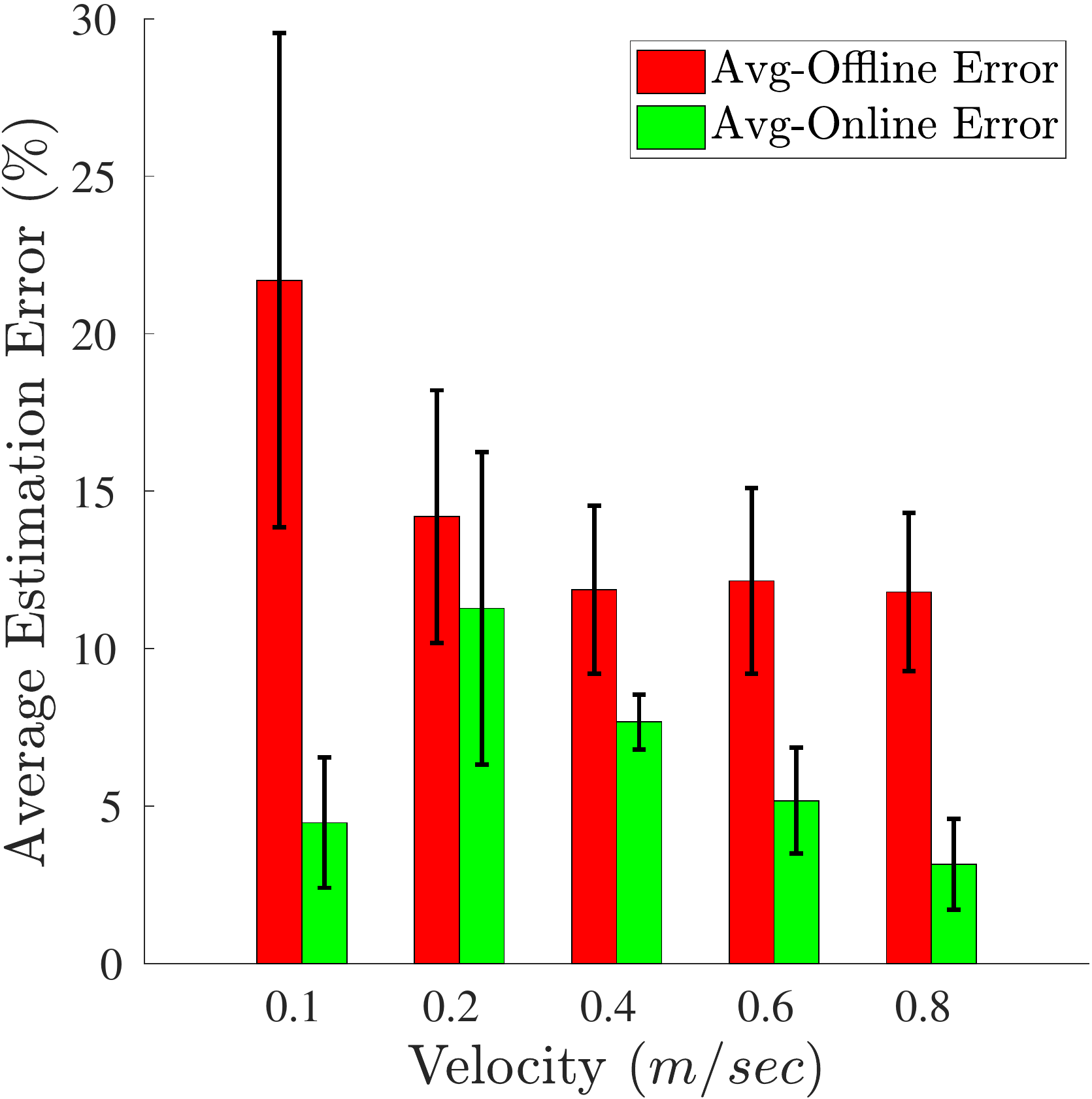}
 \caption{Range estimation error for ArDrone. Plot showing error in operational range calculated using the \textit{offline} and \textit{online} models along with corresponding standard deviation.}
 \label{fig:EstError}
 \end{figure}

Additionally, several outdoor trials were performed using the AR Drone in occasionally windy conditions. The results obtained using both the \textit{offline} and the \textit{online} variants of generalized ORangE are shown in Fig.~\ref{fig:EstError}.

Across multiple trials, the average accuracy of the \textit{offline} approach was $82.97\%$ while that of \textit{online} was $93.87\%$. For brevity, only limited results are showcased here for the generalized ORangE approach, but additional results can be found in \cite{tiwari2019unified,tiwari2018thesis}.

\section{Conclusion} 
\label{sec:conclusion}
The aim of this paper was to summarize our recent findings in \textit{operational range estimation}. Operational range estimation is an important system consideration for exploration of extreme environments such as underwater and benthic habitats, hot-springs, volcanoes, asteroids, and planetary surfaces. In most of these cases, recharging of batteries is not feasible, and, hence, the robots must be able to optimize the area explored on a single discharge cycle. To this end, we briefly presented our recent works which describe two range estimators with different levels of generality and model fidelity. Additionally, we presented arguments as to why one should use this setting as a potential mission characterization metric as opposed to the conventional \textit{energy \& endurance estimation} methods. Model performance metrics were presented to empirically validate the proposed estimators on physical robot platforms for both indoor and outdoor settings. We believe that our ORangE framework can be utilized as an important mission characteristic for applications like environmental monitoring, precision agriculture, and disaster response where robots are increasingly being deployed.

Both the simplified and generic ORangE approaches presented herewith rely on the system calibration parameters which need to be calibrated \textit{a priori}. As the robots operate in the field and undergo wear-and-tear, these parameters also vary over time. Continued monitoring of these parameters is a labor-intensive task, thus, in the future works, we would like to investigate mechanisms to reduce the reliance on such parameters. It would also be of interest to validate our generalized ORangE on other robot platforms, \textit{e.g.,} marine robots that are tasked with monitoring the marine ecology as shown in the works like \cite{dolan2007harmful,wilson2018adaptive}.

\bibliographystyle{plainnat}
\bibliography{main}

\end{document}